\documentclass{article} 
\usepackage{iclr2018_conference,times}
\usepackage{hyperref}
\usepackage{url}

\usepackage[T1]{fontenc}
\usepackage{mathrsfs}
\usepackage{mathtools}
\usepackage{array}
\usepackage{amsmath}
\usepackage{amssymb}
\usepackage{color}
\usepackage{refcount}
\usepackage[ruled,vlined]{algorithm2e} 
\usepackage{booktabs}
\usepackage{graphicx}
\usepackage{subfig}
\usepackage{siunitx}
\DeclareMathOperator{\BN}{BN}

\DeclareMathOperator*{\argmax}{arg\,max}
\def\code#1{\texttt{#1}}

\title{Normalized Direction-preserving Adam}

\author{Zijun Zhang\\
Department of Computer Science\\
University of Calgary\\
\texttt{zijun.zhang@ucalgary.ca}\\
\And
Lin Ma\\
School of Computer Science\\
Wuhan University\\
\texttt{linmawhu@gmail.com}\\
\And
Zongpeng Li\\
Department of Computer Science\\
University of Calgary\\
\texttt{zongpeng@ucalgary.ca}\\
\And
Chuan Wu\\
Department of Computer Science\\
The University of Hong Kong\\
\texttt{cwu@cs.hku.hk}\\
}

\iclrfinalcopy 

\begin{document}

\maketitle

\begin{abstract}
Adaptive optimization algorithms, such as Adam and RMSprop, have shown better optimization performance than stochastic gradient descent (SGD) in some scenarios. However, recent studies show that they often lead to worse generalization performance than SGD, especially for training deep neural networks (DNNs). In this work, we identify the reasons that Adam generalizes worse than SGD, and develop a variant of Adam to eliminate the generalization gap. The proposed method, normalized direction-preserving Adam (ND-Adam), enables more precise control of the direction and step size for updating weight vectors, leading to significantly improved generalization performance. Following a similar rationale, we further improve the generalization performance in classification tasks by regularizing the softmax logits. By bridging the gap between SGD and Adam, we also hope to shed light on why certain optimization algorithms generalize better than others.
\end{abstract}

\section{Introduction}
In contrast with the growing complexity of neural network architectures \citep{szegedy2015going,he2016deep,hu2018squeeze}, the training methods remain relatively simple. Most practical optimization methods for deep neural networks (DNNs) are based on the stochastic gradient descent (SGD) algorithm. However, the learning rate of SGD, as a hyperparameter, is often difficult to tune, since the magnitudes of different parameters vary widely, and adjustment is required throughout the training process.

To tackle this problem, several adaptive variants of SGD were developed, including Adagrad \citep{duchi2011adaptive}, Adadelta \citep{zeiler2012adadelta}, RMSprop \citep{tieleman2012rmsprop}, Adam \citep{kingma2015adam}. These algorithms aim to adapt the learning rate to different parameters automatically, based on the statistics of gradient. Although they usually simplify learning rate settings, and lead to faster convergence, it is observed that their generalization performance tend to be significantly worse than that of SGD in some scenarios \citep{wilson2017marginal}. This intriguing phenomenon may explain why SGD (possibly with momentum) is still prevalent in training state-of-the-art deep models, especially feedforward DNNs \citep{szegedy2015going,he2016deep,hu2018squeeze}. Furthermore, recent work has shown that DNNs are capable of fitting noise data \citep{zhang2017understanding}, suggesting that their generalization capabilities are not the mere result of DNNs themselves, but are entwined with optimization \citep{arpit2017closer}.

This work aims to bridge the gap between SGD and Adam in terms of the generalization performance. To this end, we identify two problems that may degrade the generalization performance of Adam, and show how these problems are (partially) avoided by using SGD with L2 weight decay. First, the updates of SGD lie in the span of historical gradients, whereas it is not the case for Adam. This difference has been discussed in rather recent literature \citep{wilson2017marginal}, where the authors show that adaptive methods can find drastically different but worse solutions than SGD. Second, while the magnitudes of Adam parameter updates are invariant to rescaling of the gradient, the effect of the updates on the {\em same} overall network function still varies with the magnitudes of parameters. As a result, the effective learning rates of weight vectors tend to decrease during training, which leads to sharp local minima that do not generalize well \citep{hochreiter1997flat}.

To address these two problems of Adam, we propose the normalized direction-preserving Adam (ND-Adam) algorithm, which controls the update direction and step size in a more precise way. We show that ND-Adam is able to achieve significantly better generalization performance than vanilla Adam, and matches that of SGD in image classification tasks.

We summarize our contributions as follows:
\begin{itemize}
	\item We observe that the directions of Adam parameter updates are different from that of SGD, i.e., Adam does not preserve the directions of gradients as SGD does. We fix the problem by adapting the learning rate to each weight vector, instead of each individual weight, such that the direction of the gradient is preserved.
	\item For both Adam and SGD without L2 weight decay, we observe that the magnitude of each vector's direction change depends on its L2-norm. We show that, using SGD with L2 weight decay implicitly normalizes the weight vectors, and thus remove the dependence in an approximate manner. We fix the problem for Adam by explicitly normalizing each weight vector, and by optimizing only its direction, such that the effective learning rate can be precisely controlled.
	\item We further demonstrate that, without proper regularization, the learning signal backpropagated from the softmax layer may vary with the overall magnitude of the logits in an undesirable way. Based on the observation, we apply batch normalization or L2-regularization to the logits, which further improves the generalization performance in classification tasks.
\end{itemize}

In essence, our proposed methods, ND-Adam and regularized softmax, improve the generalization performance of Adam by enabling more precise control over the directions of parameter updates, the learning rates, and the learning signals.

The remainder of this paper is organized as follows. In Sec.~\ref{sec:motiv}, we identify two problems of Adam, and show how SGD with L2 weight decay partially avoids these problems. In Sec.~\ref{sec:ndadam}, we further discuss and develop ND-Adam as a solution to the two problems. In Sec.~\ref{sec:bn-softmax}, we propose regularized softmax to improve the learning signal backpropagated from the softmax layer. We provide empirical evidence for our analysis, and evaluate the performance of the proposed methods in Sec.~\ref{sec:exprm}. \footnote{Code is available at \url{https://github.com/zj10/ND-Adam}.}

\section{Background and Motivation}\label{sec:motiv}
\subsection{Adaptive Moment Estimation (Adam)}\label{sec:adam}
Adaptive moment estimation (Adam) \citep{kingma2015adam} is a stochastic optimization method that applies individual adaptive learning rates to different parameters, based on the estimates of the first and second moments of the gradients. Specifically, for $n$ trainable parameters, $\theta\in\mathbb{R}^{n}$, Adam maintains a running average of the first and second moments of the gradient w.r.t. each parameter as
	\begin{subequations}
		\begin{equation}\label{eq:1st_moment}
		m_{t}=\beta_{1}m_{t-1}+\left(1-\beta_{1}\right)g_{t},
		\end{equation}
		\text{and}
		\begin{equation}\label{eq:2nd_moment}
		v_{t}=\beta_{2}v_{t-1}+\left(1-\beta_{2}\right)g_{t}^{2}.
		\end{equation}
	\end{subequations}
Here, $t$ denotes the time step, $m_{t}\in\mathbb{R}^{n}$ and $v_{t}\in\mathbb{R}^{n}$ denote respectively the first and second moments, and $\beta_{1}\in\mathbb{R}$ and $\beta_{2}\in\mathbb{R}$ are the corresponding decay factors. \cite{kingma2015adam} further notice that, since $m_{0}$ and $v_{0}$ are initialized to $0$'s, they are biased towards zero during the initial time steps, especially when the decay factors are large (i.e., close to $1$). Thus, for computing the next update, they need to be corrected as
	\begin{equation}\label{eq:crr_moment}
	\hat{m}_{t}=\frac{m_{t}}{1-\beta_{1}^{t}},\hat{v}_{t}=\frac{v_{t}}{1-\beta_{2}^{t}},
	\end{equation}
where $\beta_{1}^{t}$, $\beta_{2}^{t}$ are the $t$-th powers of $\beta_{1}$, $\beta_{2}$ respectively. Then, we can update each parameter as
	\begin{equation}\label{eq:adam_update}
	\theta_{t}=\theta_{t-1}-\frac{\alpha_{t}}{\sqrt{\hat{v}_{t}}+\epsilon}\hat{m}_{t},
	\end{equation}
where $\alpha_{t}$ is the global learning rate, and $\epsilon$ is a small constant to avoid division by zero. Note the above computations between vectors are element-wise.

A distinguishing merit of Adam is that the magnitudes of parameter updates are invariant to rescaling of the gradient, as shown by the adaptive learning rate term, $\alpha_{t}/\left(\sqrt{\hat{v}_{t}}+\epsilon\right)$. However, there are two potential problems when applying Adam to DNNs.

First, in some scenarios, DNNs trained with Adam generalize worse than that trained with stochastic gradient descent (SGD) \citep{wilson2017marginal}. \cite{zhang2017understanding} demonstrate that over-parameterized DNNs are capable of memorizing the entire dataset, no matter if it is natural data or meaningless noise data, and thus suggest much of the generalization power of DNNs comes from the training algorithm, e.g., SGD and its variants. It coincides with another recent work \citep{wilson2017marginal}, which shows that simple SGD often yields better generalization performance than adaptive gradient methods, such as Adam. As pointed out by the latter, the difference in the generalization performance may result from the different directions of updates. Specifically, for each hidden unit, the SGD update of its input weight vector can only lie in the span of all possible input vectors, which, however, is not the case for Adam due to the individually adapted learning rates. We refer to this problem as the {\em direction missing problem}.

Second, while batch normalization \citep{ioffe2015batch} can significantly accelerate the convergence of DNNs, the input weights and the scaling factor of each hidden unit can be scaled in infinitely many (but consistent) ways, without changing the function implemented by the hidden unit. Thus, for different magnitudes of an input weight vector, the updates given by Adam can have different effects on the overall network function, which is undesirable. Furthermore, even when batch normalization is not used, a network using linear rectifiers (e.g., ReLU, leaky ReLU) as activation functions, is still subject to ill-conditioning of the parameterization \citep{glorot2011deep}, and hence the same problem. We refer to this problem as the {\em ill-conditioning problem}.

\subsection{L2 Weight Decay}\label{sec:l2wd}
L2 weight decay is a regularization technique frequently used with SGD. It often has a significant effect on the generalization performance of DNNs. Despite its simplicity and crucial role in the training process, how L2 weight decay works in DNNs remains to be explained. A common justification is that L2 weight decay can be introduced by placing a Gaussian prior upon the weights, when the objective is to find the maximum {\em a posteriori} (MAP) weights \citep{blundell2015weight}. However, as discussed in Sec.~\ref{sec:adam}, the magnitudes of input weight vectors are irrelevant in terms of the overall network function, in some common scenarios, rendering the variance of the Gaussian prior meaningless.

We propose to view L2 weight decay in neural networks as a form of weight normalization, which may better explain its effect on the generalization performance. Consider a neural network trained with the following loss function:
	\begin{equation}\label{eq:loss_fcn}
	\widetilde{L}\left(\theta;\mathcal{D}\right)=L\left(\theta;\mathcal{D}\right)+\frac{\lambda}{2}\sum_{i\in\mathcal{N}}\left\Vert w_{i}\right\Vert_{2}^{2},
	\end{equation}
where $L\left(\theta;\mathcal{D}\right)$ is the original loss function specified by the task, $\mathcal{D}$ is a batch of training data, $\mathcal{N}$ is the set of all hidden units, and $w_{i}$ denotes the input weights of hidden unit $i$, which is included in the trainable parameters, $\theta$. For simplicity, we consider SGD updates without momentum. Therefore, the update of $w_{i}$ at each time step is
	\begin{equation}\label{eq:delta_w}
	\Delta w_{i}=-\alpha\frac{\partial\widetilde{L}}{\partial w_{i}}=-\alpha\left(\frac{\partial L}{\partial w_{i}}+\lambda w_{i}\right),
	\end{equation}
where $\alpha$ is the learning rate. As we can see from Eq.~\eqref{eq:delta_w}, the gradient magnitude of the L2 penalty is proportional to $\left\Vert w_{i}\right\Vert_{2}$, thus forms a negative feedback loop that stabilizes $\left\Vert w_{i}\right\Vert_{2}$ to an equilibrium value. Empirically, we find that $\left\Vert w_{i}\right\Vert_{2}$ tends to increase or decrease dramatically at the beginning of the training, and then varies mildly within a small range, which indicates $\left\Vert w_{i}\right\Vert _{2}\approx\left\Vert w_{i}+\Delta w_{i}\right\Vert _{2}$. In practice, we usually have $\left\Vert \Delta w_{i}\right\Vert _{2}/\left\Vert w_{i}\right\Vert _{2}\ll1$, thus $\Delta w_{i}$ is approximately orthogonal to $w_{i}$, i.e. $w_{i}\cdot\Delta w_{i}\approx 0$.

Let $l_{\parallel w_{i}}$ and $l_{\perp w_{i}}$ be the vector projection and rejection of $\frac{\partial L}{\partial w_{i}}$ on $w_{i}$, which are defined as
	\begin{equation}\label{eq:l_w}
	l_{\parallel w_{i}}=\left(\frac{\partial L}{\partial w_{i}}\cdot\frac{w_{i}}{\left\Vert w_{i}\right\Vert _{2}}\right)\frac{w_{i}}{\left\Vert w_{i}\right\Vert _{2}},l_{\perp w_{i}}=\frac{\partial L}{\partial w_{i}}-l_{\parallel w_{i}}.
	\end{equation}
From Eq.~\eqref{eq:delta_w} and \eqref{eq:l_w}, it is easy to show
	\begin{equation}\label{eq:delta_ratio}
	\frac{\left\Vert \Delta w_{i}\right\Vert _{2}}{\left\Vert w_{i}\right\Vert _{2}}\approx\frac{\left\Vert l_{\perp w_{i}}\right\Vert _{2}}{\left\Vert l_{\parallel w_{i}}\right\Vert _{2}}\alpha\lambda.
	\end{equation}

As discussed in Sec.~\ref{sec:adam}, when batch normalization is used, or when linear rectifiers are used as activation functions, the magnitude of $\left\Vert w_{i}\right\Vert _{2}$ becomes irrelevant; it is the direction of $w_{i}$ that actually makes a difference in the overall network function. If L2 weight decay is not applied, the magnitude of $w_{i}$'s direction change will decrease as $\left\Vert w_{i}\right\Vert _{2}$ increases during the training process, which can potentially lead to overfitting (discussed in detail in Sec.~\ref{sec:weight_optm}). On the other hand, Eq.~\eqref{eq:delta_ratio} shows that L2 weight decay implicitly normalizes the weights, such that the magnitude of $w_{i}$'s direction change does not depend on $\left\Vert w_{i}\right\Vert _{2}$, and can be tuned by the product of $\alpha$ and $\lambda$. In the following, we refer to $\left\Vert \Delta w_{i}\right\Vert _{2}/\left\Vert w_{i}\right\Vert _{2}$ as the {\em effective learning rate} of $w_{i}$.

While L2 weight decay produces the normalization effect in an implicit and approximate way, we will show that explicitly doing so enables more precise control of the effective learning rate.

\section{Normalized Direction-preserving Adam}\label{sec:ndadam}
We first present the normalized direction-preserving Adam (ND-Adam) algorithm, which essentially improves the optimization of the input weights of hidden units, while employing the vanilla Adam algorithm to update other parameters. Specifically, we divide the trainable parameters, $\theta$, into two sets, $\theta^{v}$ and $\theta^{s}$, such that $\theta^{v}=\left\{w_{i}|i\in\mathcal{N}\right\}$, and $\theta^{s}=\left\{\theta\setminus\theta^{v}\right\}$. Then we update $\theta^{v}$ and $\theta^{s}$ by different rules, as described by Alg.~\ref{alg:ND-Adam}. The learning rates for the two sets of parameters are denoted by $\alpha_{t}^{v}$ and $\alpha_{t}^{s}$, respectively.

\begin{algorithm}[tb]
		\tcc{Initialization}
		$t\gets 0$\;
		\For{$i\in \mathcal{N}$}{
			$w_{i,0}\gets w_{i,0}/\left\Vert w_{i,0}\right\Vert _{2}$\;
			$m_{0}\left(w_{i}\right)\gets 0$\;
			$v_{0}\left(w_{i}\right)\gets 0$\;
		}
		\tcc{Perform $T$ iterations of training}
		\While{$t<T$}{
			$t\gets t+1$\;
			\tcc{Update $\theta^{v}$}
			\For{$i\in \mathcal{N}$}{
				$\bar{g}_{t}\left(w_{i}\right)\gets \partial L/\partial w_{i}$\;
				$g_{t}\left(w_{i}\right)\gets\bar{g}_{t}\left(w_{i}\right)-\left(\bar{g}_{t}\left(w_{i}\right)\cdot w_{i,t-1}\right)w_{i,t-1}$\;
				$m_{t}\left(w_{i}\right)\gets\beta_{1}m_{t-1}\left(w_{i}\right)+\left(1-\beta_{1}\right)g_{t}\left(w_{i}\right)$\;
				$v_{t}\left(w_{i}\right)\gets\beta_{2}v_{t-1}\left(w_{i}\right)+\left(1-\beta_{2}\right)\left\Vert g_{t}\left(w_{i}\right)\right\Vert _{2}^{2}$\;
				$\hat{m}_{t}\left(w_{i}\right)\gets m_{t}\left(w_{i}\right)/\left(1-\beta_{1}^{t}\right)$\;
				$\hat{v}_{t}\left(w_{i}\right)\gets v_{t}\left(w_{i}\right)/\left(1-\beta_{2}^{t}\right)$\;
				$\bar{w}_{i,t}\gets w_{i,t-1}-\alpha_{t}^{v}\hat{m}_{t}\left(w_{i}\right)/\left(\sqrt{\hat{v}_{t}\left(w_{i}\right)}+\epsilon\right)$\;
				$w_{i,t}\gets\bar{w}_{i,t}/\left\Vert \bar{w}_{i,t}\right\Vert _{2}$\;
			}
			\tcc{Update $\theta^{s}$ using Adam}
			$\theta_{t}^{s}\gets\mathrm{AdamUpdate}\left(\theta_{t-1}^{s};\alpha_{t}^{s},\beta_{1},\beta_{2}\right)$\;
		}
		\Return $\theta_{T}$\;
	\caption{Normalized direction-preserving Adam \label{alg:ND-Adam}}
\end{algorithm}

In Alg.~\ref{alg:ND-Adam}, computing $g_{t}\left(w_{i}\right)$ and $w_{i,t}$ may take slightly more time compared to Adam, which however is negligible in practice. On the other hand, to estimate the second order moment of each $w_{i}\in\mathbb{R}^{n}$, Adam maintains $n$ scalars, whereas ND-Adam requires only one scalar, $v_{t}\left(w_{i}\right)$, and thus reduces the memory overhead of Adam.

In the following, we address the direction missing problem and the ill-conditioning problem discussed in Sec.~\ref{sec:adam}, and explain Alg.~\ref{alg:ND-Adam} in detail. We show how the proposed algorithm jointly solves the two problems, as well as its relation to other normalization schemes.

\subsection{Preserving Gradient Directions}
Assuming the stationarity of a hidden unit's input distribution, the SGD update (possibly with momentum) of the input weight vector is a linear combination of historical gradients, and thus can only lie in the span of the input vectors. Consequently, the input weight vector itself will eventually converge to the same subspace.

In contrast, the Adam algorithm adapts the global learning rate to each scalar parameter independently, such that the gradient of each parameter is normalized by a running average of its magnitudes, which changes the direction of the gradient. To preserve the direction of the gradient w.r.t. each input weight vector, we generalize the learning rate adaptation scheme from scalars to vectors.

Let $g_{t}\left(w_{i}\right)$, $m_{t}\left(w_{i}\right)$, $v_{t}\left(w_{i}\right)$ be the counterparts of $g_{t}$, $m_{t}$, $v_{t}$ for vector $w_{i}$. Since Eq.~\eqref{eq:1st_moment} is a linear combination of historical gradients, it can be extended to vectors without any change; or equivalently, we can rewrite it for each vector as
	\begin{equation}\label{eq:1st_moment_vector}
	m_{t}\left(w_{i}\right)=\beta_{1}m_{t-1}\left(w_{i}\right)+\left(1-\beta_{1}\right)g_{t}\left(w_{i}\right).
	\end{equation}
We then extend Eq.~\eqref{eq:2nd_moment} as
	\begin{equation}\label{eq:2nd_moment_vector}
	v_{t}\left(w_{i}\right)=\beta_{2}v_{t-1}\left(w_{i}\right)+\left(1-\beta_{2}\right)\left\Vert g_{t}\left(w_{i}\right)\right\Vert _{2}^{2},
	\end{equation}
i.e., instead of estimating the average gradient magnitude for each individual parameter, we estimate the average of $\left\Vert g_{t}\left(w_{i}\right)\right\Vert _{2}^{2}$ for each vector $w_{i}$. In addition, we modify Eq.~\eqref{eq:crr_moment} and \eqref{eq:adam_update} accordingly as
	\begin{equation}\label{eq:crr_moment_vector}
	\hat{m}_{t}\left(w_{i}\right)=\frac{m_{t}\left(w_{i}\right)}{1-\beta_{1}^{t}},\hat{v}_{t}\left(w_{i}\right)=\frac{v_{t}\left(w_{i}\right)}{1-\beta_{2}^{t}},
	\end{equation}
and
	\begin{equation}\label{eq:ndadam_update_vector}
	w_{i,t}=w_{i,t-1}-\frac{\alpha_{t}^{v}}{\sqrt{\hat{v}_{t}\left(w_{i}\right)}+\epsilon}\hat{m}_{t}\left(w_{i}\right).
	\end{equation}
Here, $\hat{m}_{t}\left(w_{i}\right)$ is a vector with the same dimension as $w_{i}$, whereas $\hat{v}_{t}\left(w_{i}\right)$ is a scalar. Therefore, when applying Eq.~\eqref{eq:ndadam_update_vector}, the direction of the update is the negative direction of $\hat{m}_{t}\left(w_{i}\right)$, and thus is in the span of the historical gradients of $w_{i}$.

Despite the empirical success of SGD, a question remains as to why it is desirable to constrain the input weights in the span of the input vectors. A possible explanation is related to the manifold hypothesis, which suggests that real-world data presented in high dimensional spaces (e.g., images, audios, text) concentrates on manifolds of much lower dimensionality \citep{cayton2005algorithms,narayanan2010sample}. In fact, commonly used activation functions, such as (leaky) ReLU, sigmoid, tanh, can only be activated (not saturating or having small gradients) by a portion of the input vectors, in whose span the input weights lie upon convergence. Assuming the local linearity of the manifolds of data or hidden-layer representations, constraining the input weights in the subspace that contains that portion of the input vectors, encourages the hidden units to form local coordinate systems on the corresponding manifold, which can lead to good representations \citep{rifai2011manifold}.

\subsection{Spherical Weight Optimization}\label{sec:weight_optm}
The ill-conditioning problem occurs when the magnitude change of an input weight vector can be compensated by other parameters, such as the scaling factor of batch normalization, or the output weight vector, without affecting the overall network function. Consequently, suppose we have two DNNs that parameterize the same function, but with some of the input weight vectors having different magnitudes, applying the same SGD or Adam update rule will, in general, change the network functions in different ways. Thus, the ill-conditioning problem makes the training process inconsistent and difficult to control.

More importantly, when the weights are not properly regularized (e.g., without using L2 weight decay), the magnitude of $w_{i}$'s direction change will decrease as $\left\Vert w_{i}\right\Vert _{2}$ increases during the training process. As a result, the effective learning rate for $w_{i}$ tends to decrease faster than expected. The gradient noise introduced by large learning rates is crucial to avoid sharp minima \citep{smith2018bayesian}. And it is well known that sharp minima generalize worse than flat minima \citep{hochreiter1997flat}.

As shown in Sec.~\ref{sec:l2wd}, when combined with SGD, L2 weight decay can alleviate the ill-conditioning problem by implicitly and approximately normalizing the weights. However, the approximation fails when $\left\Vert w_{i}\right\Vert _{2}$ is far from the equilibrium due to improper initialization, or drastic changes in the magnitudes of the weight vectors. In addition, due to the direction missing problem, naively applying L2 weight decay to Adam does not yield the same effect as it does on SGD. In concurrent work, \cite{loshchilov2017fixing} address the problem by decoupling the weight decay and the optimization steps taken w.r.t. the loss function. However, their experimental results indicate that improving L2 weight decay alone cannot eliminate the generalization gap between Adam and SGD.

The ill-conditioning problem is also addressed by \cite{neyshabur2015path}, by employing a geometry invariant to rescaling of weights. However, their proposed methods do not preserve the direction of gradient.

To address the ill-conditioning problem in a more principled way, we restrict the L2-norm of each $w_{i}$ to $1$, and only optimize its direction. In other words, instead of optimizing $w_{i}$ in a $n$-dimensional space, we optimize $w_{i}$ on a $\left(n-1\right)$-dimensional unit sphere. Specifically, we first compute the raw gradient w.r.t. $w_{i}$, $\bar{g}_{t}\left(w_{i}\right)=\partial L/\partial w_{i}$, and project the gradient onto the unit sphere as
	\begin{equation}\label{eq:sphr_grad}
	g_{t}\left(w_{i}\right)=\bar{g}_{t}\left(w_{i}\right)-\left(\bar{g}_{t}\left(w_{i}\right)\cdot w_{i,t-1}\right)w_{i,t-1}.
	\end{equation}
Here, $\left\Vert w_{i,t-1}\right\Vert _{2}=1$. Then we follow Eq.~\eqref{eq:1st_moment_vector}-\eqref{eq:crr_moment_vector}, and replace \eqref{eq:ndadam_update_vector} with
	\begin{equation}\label{eq:ndadam_update_normed}
	\bar{w}_{i,t}=w_{i,t-1}-\frac{\alpha_{t}^{v}}{\sqrt{\hat{v}_{t}\left(w_{i}\right)}+\epsilon}\hat{m}_{t}\left(w_{i}\right),
	\text{ and }
	w_{i,t}=\frac{\bar{w}_{i,t}}{\left\Vert \bar{w}_{i,t}\right\Vert _{2}}.
	\end{equation}

In Eq.~\eqref{eq:sphr_grad}, we keep only the component that is orthogonal to $w_{i,t-1}$. However, $\hat{m}_{t}\left(w_{i}\right)$ is not necessarily orthogonal as well; moreover, even when $\hat{m}_{t}\left(w_{i}\right)$ is orthogonal to $w_{i,t-1}$, $\left\Vert w_{i}\right\Vert _{2}$ can still increase according to the Pythagorean theorem. Therefore, we explicitly normalize $w_{i,t}$ in Eq.~\eqref{eq:ndadam_update_normed}, to ensure $\left\Vert w_{i,t}\right\Vert _{2}=1$ after each update. Also note that, since $w_{i,t-1}$ is a linear combination of its historical gradients, $g_{t}\left(w_{i}\right)$ still lies in the span of the historical gradients after the projection in Eq.~\eqref{eq:sphr_grad}.

Compared to SGD with L2 weight decay, spherical weight optimization explicitly normalizes the weight vectors, such that each update to the weight vectors only changes their directions, and strictly keeps the magnitudes constant. As a result, the effective learning rate of a weight vector is
	\begin{equation}
	\frac{\left\Vert \Delta w_{i,t}\right\Vert _{2}}{\left\Vert w_{i,t-1}\right\Vert _{2}}\approx\frac{\left\Vert \hat{m}_{t}\left(w_{i}\right)\right\Vert _{2}}{\sqrt{\hat{v}_{t}\left(w_{i}\right)}}\alpha_{t}^{v},
	\end{equation}
which enables precise control over the learning rate of $w_{i}$ through a single hyperparameter, $\alpha_{t}^{v}$, rather than two as required by Eq.~\eqref{eq:delta_ratio}.

Note that it is possible to control the effective learning rate more precisely, by normalizing $\hat{m}_{t}\left(w_{i}\right)$ with $\left\Vert \hat{m}_{t}\left(w_{i}\right)\right\Vert _{2}$, instead of by $\sqrt{\hat{v}_{t}\left(w_{i}\right)}$. However, by doing so, we lose information provided by $\left\Vert \hat{m}_{t}\left(w_{i}\right)\right\Vert _{2}$ at different time steps. In addition, since $\hat{m}_{t}\left(w_{i}\right)$ is less noisy than $g_{t}\left(w_{i}\right)$, $\left\Vert \hat{m}_{t}\left(w_{i}\right)\right\Vert _{2}/\sqrt{\hat{v}_{t}\left(w_{i}\right)}$ becomes small near convergence, which is considered a desirable property of Adam \citep{kingma2015adam}. Thus, we keep the gradient normalization scheme intact.

We note the difference between various gradient normalization schemes and the normalization scheme employed by spherical weight optimization. As shown in Eq.~\eqref{eq:ndadam_update_vector}, ND-Adam generalizes the gradient normalization scheme of Adam, and thus both Adam and ND-Adam normalize the gradient by a running average of its magnitude. This, and other similar schemes \citep{hazan2015beyond,yu2017normalized} make the optimization less susceptible to vanishing and exploding gradients. The proposed spherical weight optimization serves a different purpose. It normalizes each weight vector and projects the gradient onto a unit sphere, such that the effective learning rate can be controlled more precisely. Moreover, it provides robustness to improper weight initialization, since the magnitude of each weight vector is kept constant.

For nonlinear activation functions (without batch normalization), such as sigmoid and tanh, an extra scaling factor is needed for each hidden unit to express functions that require unnormalized weight vectors. For instance, given an input vector $x\in\mathbb{R}^{n}$, and a nonlinearity $\phi\left(\cdot\right)$, the activation of hidden unit $i$ is then given by
	\begin{equation}\label{eq:hu_swo}
	y_{i}=\phi\left(\gamma_{i} w_{i}\cdot x+b_{i}\right),
	\end{equation}
where $\gamma_{i}$ is the scaling factor, and $b_{i}$ is the bias. Consequently, normalizing weight vectors does not limit the expressiveness of models.

\subsection{Relation to Weight Normalization and Batch Normalization}
A related normalization and reparameterization scheme, weight normalization \citep{salimans2016weight}, has been developed as an alternative to batch normalization, aiming to accelerate the convergence of SGD optimization. We note the difference between spherical weight optimization and weight normalization. First, the weight vector of each hidden unit is not directly normalized in weight normalization, i.e, $\left\Vert w_{i}\right\Vert _{2}\neq 1$ in general. At training time, the activation of hidden unit $i$ is
	\begin{equation}
	y_{i}=\phi\left(\frac{\gamma_{i}}{\left\Vert w_{i}\right\Vert _{2}}w_{i}\cdot x+b_{i}\right),
	\end{equation}
which is equivalent to Eq.~\eqref{eq:hu_swo} for the forward pass. For the backward pass, the effective learning rate still depends on $\left\Vert w_{i}\right\Vert _{2}$ in weight normalization, hence it does not solve the ill-conditioning problem. At inference time, both of these two schemes can merge $w_{i}$ and $\gamma_{i}$ into a single equivalent weight vector, $w'_{i}=\gamma_{i} w_{i}$, or $w'_{i}=\frac{\gamma_{i}}{\left\Vert w_{i}\right\Vert _{2}}w_{i}$.

While spherical weight optimization naturally encompasses weight normalization, it can further benefit from batch normalization. When combined with batch normalization, Eq.~\eqref{eq:hu_swo} evolves into
	\begin{equation}
	y_{i}=\phi\left(\gamma_{i}\BN\left(w_{i}\cdot x\right)+b_{i}\right), 
	\end{equation}
where $\BN\left(\cdot\right)$ represents the transformation done by batch normalization without scaling and shifting. Here, $\gamma_{i}$ serves as the scaling factor for both the normalized weight vector and batch normalization.

\section{Regularized Softmax}\label{sec:bn-softmax}
For multi-class classification tasks, the softmax function is the {\em de facto} activation function for the output layer. Despite its simplicity and intuitive probabilistic interpretation, we observe a related problem to the ill-conditioning problem we have addressed. Similar to how different magnitudes of weight vectors result in different updates to the same network function, the learning signal backpropagated from the softmax layer varies with the overall magnitude of the logits. 

Specifically, when using cross entropy as the surrogate loss with one-hot target vectors, the prediction is considered correct as long as $\argmax_{c\in\mathcal{C}}\left(z_{c}\right)$ is the target class, where $z_{c}$ is the logit before the softmax activation, corresponding to category $c\in\mathcal{C}$. Thus, the logits can be positively scaled together without changing the predictions, whereas the cross entropy and its derivatives will vary with the scaling factor. Concretely, denoting the scaling factor by $\eta$, the gradient w.r.t. each logit is
	\begin{equation}\label{eq:softmax_grad}
	\frac{\partial L}{\partial z_{\hat{c}}}=\eta\left[\frac{\exp\left(\eta z_{\hat{c}}\right)}{\sum_{c\in\mathcal{C}}\exp\left(\eta z_{c}\right)}-1\right],
	\text{ and }
	\frac{\partial L}{\partial z_{\bar{c}}}=\frac{\eta\exp\left(\eta z_{\bar{c}}\right)}{\sum_{c\in\mathcal{C}}\exp\left(\eta z_{c}\right)},
	\end{equation}
where $\hat{c}$ is the target class, and $\bar{c}\in\mathcal{C}\backslash\left\{\hat{c}\right\}$.

For Adam and ND-Adam, since the gradient w.r.t. each scalar or vector are normalized, the absolute magnitudes of Eq.~\eqref{eq:softmax_grad} are irrelevant. Instead, the relative magnitudes make a difference here. When $\eta$ is small, we have
	\begin{equation}\label{eq:rel_softmax_grad_smallLogit}
	\lim_{\eta\rightarrow0}\left|\frac{\partial L/\partial z_{\bar{c}}}{\partial L/\partial z_{\hat{c}}}\right|=\frac{1}{\left|\mathcal{C}\right|-1},
	\end{equation}
which indicates that, when the magnitude of the logits is small, softmax encourages the logit of the target class to increase, while equally penalizing that of the other classes, regardless of the difference in $\hat{z}-\bar{z}$ for different $\bar{z}\in\mathcal{C}\backslash\left\{\hat{z}\right\}$. However, it is more reasonable to penalize more the logits that are closer to $\hat{z}$, which are more likely to cause misclassification.

On the other end of the spectrum, assuming no two digits are the same, we have
	\begin{equation}\label{eq:rel_softmax_grad_largeLogit}
	\lim_{\eta\rightarrow\infty}\left|\frac{\partial L/\partial z_{\bar{c}'}}{\partial L/\partial z_{\hat{c}}}\right|=1,\lim_{\eta\rightarrow\infty}\left|\frac{\partial L/\partial z_{\bar{c}''}}{\partial L/\partial z_{\hat{c}}}\right|=0,
	\end{equation}
where $\bar{c}'=\argmax_{c\in\mathcal{C}\backslash \left\{\hat{c}\right\}}\left(z_{c}\right)$, and $\bar{c}''\in\mathcal{C}\backslash \left\{\hat{c},\bar{c}'\right\}$. Eq.~\eqref{eq:rel_softmax_grad_largeLogit} indicates that, when the magnitude of the logits is large, softmax penalizes only the largest logit of the non-target classes. In this case, although the logit that is most likely to cause misclassification is strongly penalized, the logits of other non-target classes are ignored. As a result, the logits of the non-target classes tend to be similar at convergence, ignoring the fact that some classes are closer to each other than the others. The latter case is related to the saturation problem of softmax discussed in the literature \citep{oland2017careful}, where they focus on the problem of small absolute gradient magnitude, which nevertheless does not affect Adam and ND-Adam.

We propose two methods to exploit the prior knowledge that the magnitude of the logits should not be too small or too large. First, we can apply batch normalization to the logits. But instead of setting $\gamma_{c}$'s as trainable variables, we consider them as a single hyperparameter, $\gamma_{\mathcal{C}}$, such that $\gamma_{c}=\gamma_{\mathcal{C}},\forall c\in\mathcal{C}$. Tuning the value of $\gamma_{\mathcal{C}}$ can lead to a better trade-off between the two extremes described by Eq.~\eqref{eq:rel_softmax_grad_smallLogit} and \eqref{eq:rel_softmax_grad_largeLogit}. We observe in practice that the optimal value of $\gamma_{\mathcal{C}}$ tends to be the same for different optimizers or different network widths, but varies with network depth. We refer to this method as batch-normalized softmax (BN-Softmax).

Alternatively, since the magnitude of the logits tends to grow larger than expected (in order to minimize the cross entropy), we can apply L2-regularization to the logits by adding the following penalty to the loss function:
	\begin{equation}\label{eq:l2logit}
	L_{\mathcal{C}}=\frac{\lambda_{\mathcal{C}}}{2}\sum_{c\in\mathcal{C}}z_{c}^{2},
	\end{equation}
where $\lambda_{\mathcal{C}}$ is a hyperparameter to be tuned. Different from BN-Softmax, $\lambda_{\mathcal{C}}$ can also be shared by different networks of different depths. 

\section{Experiments}\label{sec:exprm}
In this section, we provide empirical evidence for the analysis in Sec.~\ref{sec:l2wd}, and evaluate the performance of ND-Adam and regularized softmax on CIFAR-10 and CIFAR-100.

\subsection{The Effect of L2 Weight Decay}\label{sec:eff_l2wd}
To empirically examine the effect of L2 weight decay, we train a wide residual network (WRN) \citep{zagoruyko2016wide} of $22$ layers, with a width of $7.5$ times that of a vanilla ResNet. Using the notation suggested by \cite{zagoruyko2016wide}, we refer to this network as WRN-$22$-$7.5$. We train the network on the CIFAR-10 dataset \citep{krizhevsky2009learning}, with a small modification to the original WRN architecture, and with a different learning rate annealing schedule. Specifically, for simplicity and slightly better performance, we replace the last fully connected layer with a convolutional layer with $10$ output feature maps. i.e., we change the layers after the last residual block from \code{BN-ReLU-GlobalAvgPool-FC-Softmax} to \code{BN-ReLU-Conv-GlobalAvgPool-Softmax}. In addition, for clearer comparisons, the learning rate is annealed according to a cosine function without restart \citep{loshchilov2017sgdr,gastaldi2017shake}. We train the model for $80$k iterations with a batch size of $128$, similar to the settings used by \citeauthor{zagoruyko2016wide} \citep{zagoruyko2016wide}. The experiments are based on a TensorFlow implementation of WRN \citep{tensorflow2016wrn}.

\begin{figure}[h]
	\subfloat[Scalar projection of $\Delta w_{i}^{l}$ on $\Delta w_{i}^{p}$ normalized by $\left\Vert \Delta w_{i}^{p}\right\Vert _{2}$.]{
		\includegraphics[width=0.48\textwidth]{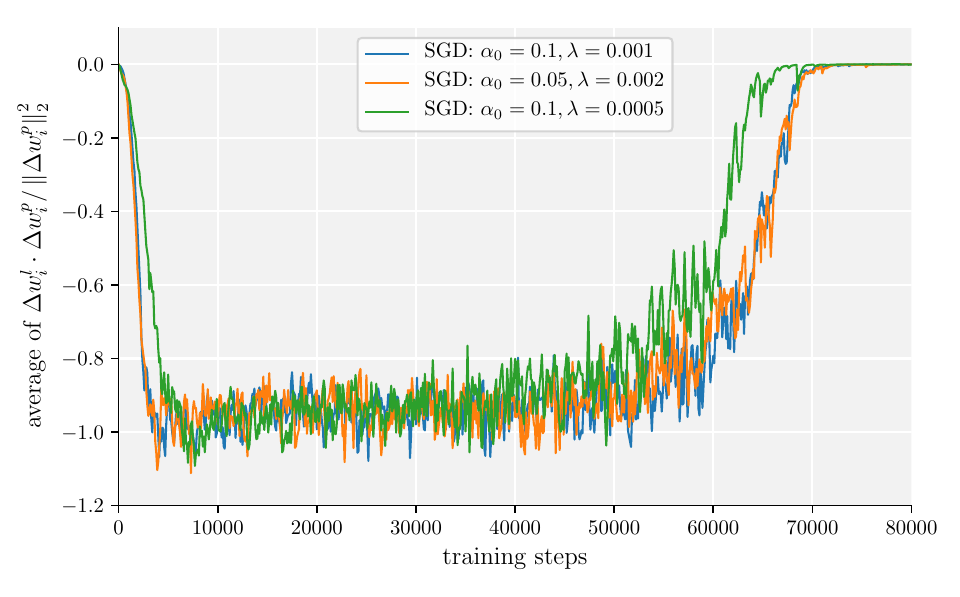}
		\label{fig:proj_ratio}}
	\hfill
	\subfloat[Relative magnitudes of weight updates, or effective learning rates.]{
		\includegraphics[width=0.48\textwidth]{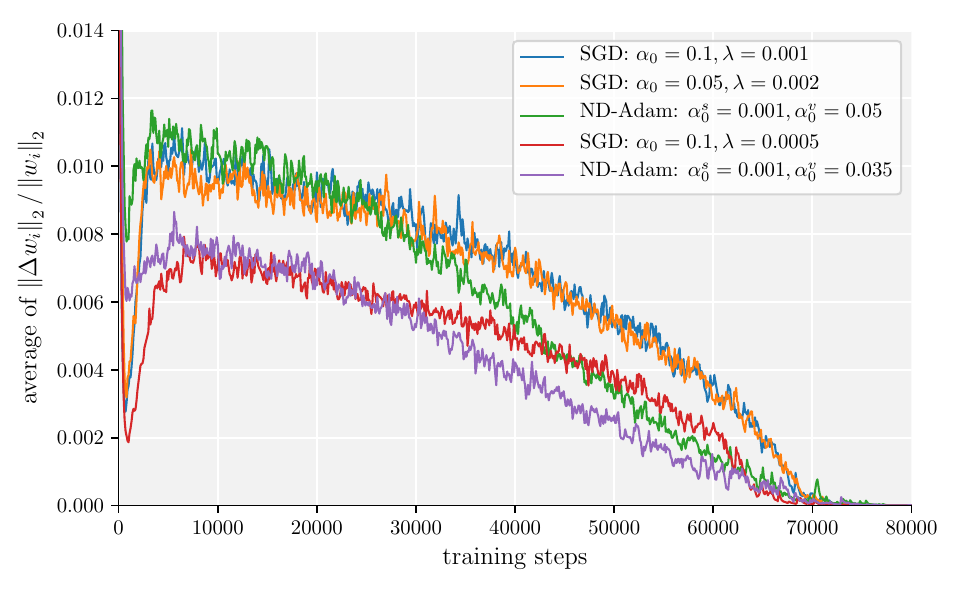}
		\label{fig:dw_w}}
	\caption{An illustration of how L2 weight decay and ND-Adam control the effective learning rate. The results are obtained from the $5$th layer of the network, and other layers show similar results.}
	\label{fig:lr_ctr}
\end{figure}

As a common practice, we use SGD with a momentum of $0.9$, the analysis for which is similar to that in Sec.~\ref{sec:l2wd}. Due to the linearity of derivatives and momentum, $\Delta w_{i}$ can be decomposed as $\Delta w_{i}=\Delta w_{i}^{l}+\Delta w_{i}^{p}$, where $\Delta w_{i}^{l}$ and $\Delta w_{i}^{p}$ are the components corresponding to the original loss function, $L\left(\cdot\right)$, and the L2 penalty term (see Eq.~\eqref{eq:loss_fcn}), respectively. Fig.~\ref{fig:proj_ratio} shows the ratio between the scalar projection of $\Delta w_{i}^{l}$ on $\Delta w_{i}^{p}$ and $\left\Vert \Delta w_{i}^{p}\right\Vert _{2}$, which indicates how the tendency of $\Delta w_{i}^{l}$ to increase $\left\Vert w_{i}\right\Vert _{2}$ is compensated by $\Delta w_{i}^{p}$. Note that $\Delta w_{i}^{p}$ points to the negative direction of $w_{i}$, even when momentum is used, since the direction change of $w_{i}$ is slow. As shown in Fig.~\ref{fig:proj_ratio}, at the beginning of the training, $\Delta w_{i}^{p}$ dominants and quickly adjusts $\left\Vert w_{i}\right\Vert _{2}$ to its equilibrium value. During the middle stage of the training, the projection of $\Delta w_{i}^{l}$ on $\Delta w_{i}^{p}$, and $\Delta w_{i}^{p}$ almost cancel each other. Then, towards the end of the training, the gradient of $w_{i}$ diminishes rapidly, making $\Delta w_{i}^{p}$ dominant again. Therefore, Eq.~\eqref{eq:delta_ratio} holds more accurately during the middle stage of the training.

In Fig.~\ref{fig:dw_w}, we show how the effective learning rate varies in different hyperparameter settings. By Eq.~\eqref{eq:delta_ratio}, $\left\Vert \Delta w_{i}\right\Vert _{2}/\left\Vert w_{i}\right\Vert _{2}$ is expected to remain the same as long as $\alpha\lambda$ stays constant, which is confirmed by the fact that the curve for $\alpha_{0}=0.1, \lambda=0.001$ overlaps with that for $\alpha_{0}=0.05, \lambda=0.002$. However, comparing the curve for $\alpha_{0}=0.1, \lambda=0.001$, with that for $\alpha_{0}=0.1, \lambda=0.0005$, we can see that the value of $\left\Vert \Delta w_{i}\right\Vert _{2}/\left\Vert w_{i}\right\Vert _{2}$ does not change proportionally to $\alpha\lambda$. On the other hand, by using ND-Adam, we can control the value of $\left\Vert \Delta w_{i}\right\Vert _{2}/\left\Vert w_{i}\right\Vert _{2}$ more precisely by adjusting the learning rate for weight vectors, $\alpha^{v}$. For the same training step, changes in $\alpha^{v}$ lead to approximately proportional changes in $\left\Vert \Delta w_{i}\right\Vert _{2}/\left\Vert w_{i}\right\Vert _{2}$, as shown by the two curves corresponding to ND-Adam in Fig.~\ref{fig:dw_w}.

\subsection{Performance Evaluation}
To compare the generalization performance of SGD, Adam, and ND-Adam, we train the same WRN-$22$-$7.5$ network on the CIFAR-10 and CIFAR-100 datasets. For SGD and ND-Adam, we first tune the hyperparameters for SGD ($\alpha_{0}=0.1, \lambda=0.001$, momentum $0.9$), then tune the initial learning rate of ND-Adam for weight vectors to match the effective learning rate to that of SGD, i.e., $\alpha_{0}^{v}=0.05$, as shown in Fig.~\ref{fig:dw_w}. While L2 weight decay can greatly affect the performance of SGD, it does not noticeably benefit Adam in our experiments. For Adam and ND-Adam, $\beta_{1}$ and $\beta_{2}$ are set to the default values of Adam, i.e., $\beta_{1}=0.9$, $\beta_{2}=0.999$. Although the learning rate of Adam is usually set to a constant value, we observe better performance with the cosine learning rate schedule. The initial learning rate of Adam ($\alpha_{0}$), and that of ND-Adam for scalar parameters ($\alpha_{0}^{s}$) are both tuned to $0.001$. We use horizontal flips and random crops for data augmentation, and no dropout is used.

\begin{figure}[h]
	\centering
	\begin{minipage}{.49\textwidth}
		\includegraphics[width=1\textwidth]{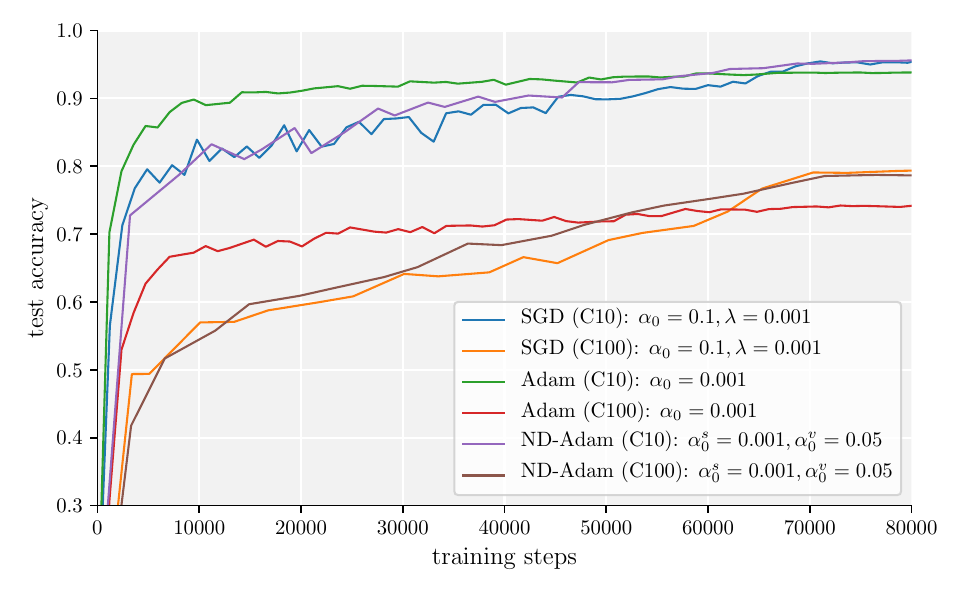}
		\caption{Test accuracies of the same network trained with SGD, Adam, and ND-Adam. Details are shown in the first part of Table~\ref{tab:test_err}.}
		\label{fig:test_accr}
	\end{minipage}
	\hfill
	\begin{minipage}{.49\textwidth}
		\includegraphics[width=1\textwidth]{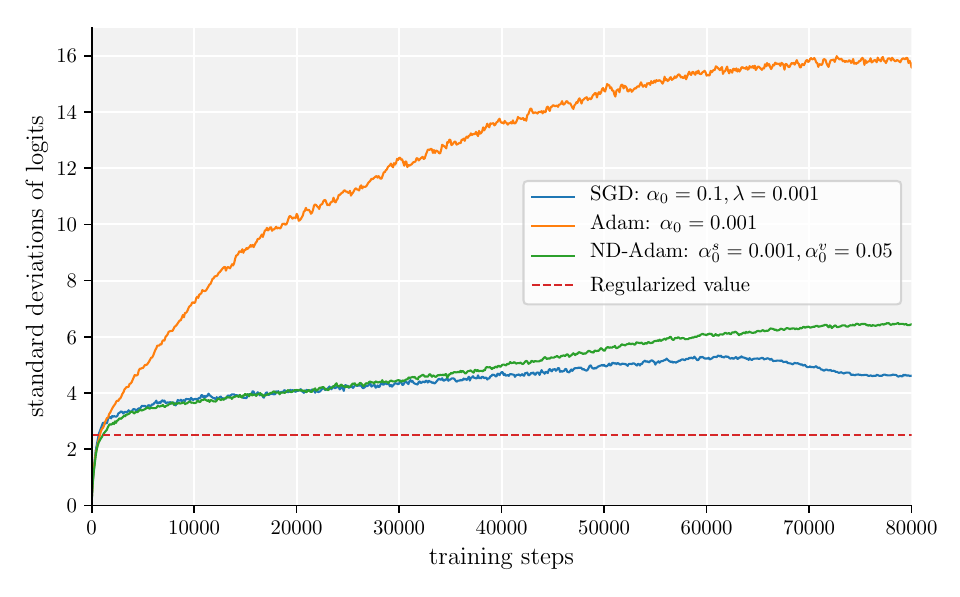}
		\caption{Magnitudes of softmax logits in different settings. Results of WRN-$22$-$7.5$ networks trained on CIFAR-10.}
		\label{fig:logit_scale}
	\end{minipage}
\end{figure}

We first experiment with the use of trainable scaling parameters ($\gamma_{i}$) of batch normalization. As shown in Fig.~\ref{fig:test_accr}, at convergence, the test accuracies of ND-Adam are significantly improved upon that of vanilla Adam, and matches that of SGD. Note that at the early stage of training, the test accuracies of Adam increase more rapidly than that of ND-Adam and SGD. However, the test accuracies remain at a high level afterwards, which indicates that Adam tends to quickly find and get stuck in bad local minima that do not generalize well.

The average results of 3 runs are summarized in the first part of Table~\ref{tab:test_err}. Interestingly, compared to SGD, ND-Adam shows slightly better performance on CIFAR-10, but worse performance on CIFAR-100. This inconsistency may be related to the problem of softmax discussed in Sec.~\ref{sec:bn-softmax}, that there is a lack of proper control over the magnitude of the logits. But overall, given comparable effective learning rates, ND-Adam and SGD show similar generalization performance. In this sense, the effective learning rate is a more natural learning rate measure than the learning rate hyperparameter.

Next, we repeat the experiments with the use of BN-Softmax. As discussed in Sec.~\ref{sec:weight_optm}, $\gamma_{i}$'s can be removed from a linear rectifier network, without changing the overall network function. Although this property does not strictly hold for residual networks due to the skip connections, we observe that when BN-Softmax is used, simply removing the scaling factors results in slightly better performance for all three algorithms. Thus, we only report results for this setting. The scaling factor of the logits, $\gamma_{\mathcal{C}}$, is set to $2.5$ for CIFAR-10, and $1$ for CIFAR-100.

As shown in the second part of Table~\ref{tab:test_err}, while we obtain the best generalization performance with ND-Adam, the improvement is most prominent for Adam, and is relatively small for SGD. This discrepancy can be explained by comparing the magnitudes of softmax logits without regularization. As shown in Fig.~\ref{fig:logit_scale}, the magnitude of logits corresponding to Adam is much larger than that of ND-Adam and SGD, and therefore benefits more from the regularization.
\begin{table}[h]
	\begin{minipage}{.48\linewidth}
		\protect\caption{Test error rates of WRN-$22$-$7.5$ networks on CIFAR-10 and CIFAR-100. Based on a TensorFlow implementation of WRN.}
		\begin{centering}
			\begin{tabular}{m{2.3cm}>{\centering}m{1.48cm}>{\centering}m{1.66cm}}
				\toprule 
				Method & CIFAR-10 Error (\%)& CIFAR-100 Error (\%)\tabularnewline
				\midrule
				\multicolumn{3}{c}{BN w/ scaling factors}\tabularnewline
				\midrule
				SGD & 4.61 & 20.60\tabularnewline
				Adam & 6.14 & 25.51\tabularnewline
				ND-Adam & 4.53 & 21.45\tabularnewline
				\midrule
				\multicolumn{3}{c}{BN w/o scaling factors, BN-Softmax}\tabularnewline
				\midrule
				SGD & 4.49 & 20.18\tabularnewline
				Adam & 5.43 & 22.48\tabularnewline
				ND-Adam & \textbf{4.14} & \textbf{19.90}\tabularnewline
				\bottomrule
			\end{tabular}
			\par\end{centering}
		\label{tab:test_err}
	\end{minipage}
	\hfill
	\begin{minipage}{.48\linewidth}
		\protect\caption{Test error rates of WRN-$22$-$7.5$ and WRN-$28$-$10$ networks on CIFAR-10 and CIFAR-100. Based on the original implementation of WRN.}
		\begin{centering}
			\begin{tabular}{m{2.3cm}>{\centering}m{1.48cm}>{\centering}m{1.66cm}}
				\toprule 
				Method & CIFAR-10 Error (\%)& CIFAR-100 Error (\%)\tabularnewline
				\midrule
				\multicolumn{3}{c}{WRN-$22$-$7.5$}\tabularnewline
				\midrule
				SGD & 3.84 & \textbf{19.24}\tabularnewline
				ND-Adam & \textbf{3.70} & 19.30\tabularnewline
				\midrule
				\multicolumn{3}{c}{WRN-$28$-$10$}\tabularnewline
				\midrule
				SGD & 3.80 & 18.48\tabularnewline
				ND-Adam & \textbf{3.70} & \textbf{18.42}\tabularnewline
				\bottomrule
			\end{tabular}
			\par\end{centering}
		\label{tab:test_err_pt}
	\end{minipage}
\end{table}

While the TensorFlow implementation we use already provides an adequate test bed, we notice that it is different from the original implementation of WRN in several aspects. For instance, they use different nonlinearities (leaky ReLU vs. ReLU), and use different skip connections for downsampling (average pooling vs. strided convolution). A subtle yet important difference is that, L2-regularization is applied not only to weight vectors, but also to the scales and biases of batch normalization in the original implementation, which leads to better generalization performance. For further comparison between SGD and ND-Adam, we reimplement ND-Adam and test its performance on a PyTorch version of the original implementation \citep{pytorch2016wrn}.

Due to the aforementioned differences, we use a slightly different hyperparameter setting in this experiment. Specifically, for SGD $\lambda$ is set to \num{5e-4}, while for ND-Adam $\lambda$ is set to \num{5e-6} (L2-regularization for biases), and both $\alpha_{0}^{s}$ and $\alpha_{0}^{v}$ are set to $0.04$. In this case, regularizing softmax does not yield improved performance for SGD, since the L2-regularization applied to $\gamma_{i}$'s and the last layer weights can serve a similar purpose. Thus, we only apply L2-regularized softmax for ND-Adam with $\lambda_{\mathcal{C}}=0.001$. The average results of 3 runs are summarized in Table~\ref{tab:test_err_pt}. Note that the performance of SGD for WRN-$28$-$10$ is slightly better than that reported with the original implementation (i.e., $4.00$ and $19.25$), due to the modifications described in Sec.~\ref{sec:eff_l2wd}. In this experiment, SGD and ND-Adam show almost identical generalization performance.

\section{Conclusion}
We introduced ND-Adam, a tailored version of Adam for training DNNs, to bridge the generalization gap between Adam and SGD. ND-Adam is designed to preserve the direction of gradient for each weight vector, and produce the regularization effect of L2 weight decay in a more precise and principled way. We further introduced regularized softmax, which limits the magnitude of softmax logits to provide better learning signals. Combining ND-Adam and regularized softmax, we show through experiments significantly improved generalization performance, eliminating the gap between Adam and SGD. From a high-level view, our analysis and empirical results suggest the need for more precise control over the training process of DNNs.

%

\bibliography{references}
\bibliographystyle{iclr2018_conference}

\end{document}